\documentclass[12pt]{article}
\usepackage{graphicx} 
\usepackage{amsmath}
\usepackage{tikz}
\usepackage{booktabs}
\usepackage{longtable}
\usepackage{pdflscape}
\usepackage{adjustbox}
\usepackage{rotating}
\usepackage{url}
\usepackage{hyperref}
\usepackage{geometry}
\usepackage{array}
\usepackage{comment}
\usepackage{listings}
\usepackage{xcolor}
\usetikzlibrary{shapes.geometric, arrows}

\title{An Open-Source Web-Based Tool for Evaluating Open-Source Large Language Models Leveraging Information Retrieval from Custom Documents}
\author{Godfrey Inyama}
\date{February 2025}

\begin{document}

\maketitle

\begin{abstract}
In our work, we present the first-of-its-kind open-source web-based tool which is able to demonstrate the impacts of a user's speech act during discourse with conversational agents, which leverages open-source large language models. With this software resource, it is possible for researchers and experts to evaluate the performance of various dialogues, visualize the user's communicative intents, and utilise uploaded specific documents for the chat agent to use for its information retrieval to respond to the user query. Almost all modern conversational agents and systems leverage the encoder-decoder architecture of generative models to first understand and generate responses to user queries. The context gathered by these models is obtained from a set of linguistic features extracted, which forms the context embeddings of the models. Regardless of these models showing good context understanding based on these features, there still remains a gap in including deeper pragmatic features to improve the model's comprehension of the query, hence the efforts to develop this web resource, which is able to extract and then inject this overlooked feature in the encoder-decoder pipeline of the conversational agent. To demonstrate the effect and impact of the resource, we carried out an experiment which evaluated the system using 2 knowledge files for information retrieval, with 2 user queries each, across 5 open-source large language models using 10 standard metrics. Our results showed that larger open-source models, such as (e.g., Llama2:13B, Llama3-ChatQA-Latest) demonstrated an improved alignment when the user speech act was included with their query, achieving higher Question-Answer and linguistic similarity scores to the source document. The smaller models in contrast (e.g., TinyLlama:Latest) showed an increased perplexity and mixed performance, which explicitly indicated struggles in processing queries that explicitly included speech act. The results from the analysis using the developed web resource highlight the potential of speech acts towards enhancing conversational depths while underscoring the need for model-specific optimizations to address increased computational costs and response times.
\end{abstract}

\maketitle

\section{Introduction}
The spontaneous evolution and transformation of conversational agents, ignited by the improvements in LLMs has transformed the way in which humans interact and respond with technology. Due to factors like adaptability, accessibility, and performance, open-source LLMs have gained significant attention. Regardless of these strengths, these open-source models often struggle with refined linguistic features such as speech acts which forms the the intended speech act behind a statement or locution, creating a gap in better understanding the speaker or user's communicative intent. Handling this limitation is essential for improving the relevance, coherence, and responsiveness of conversational agents.

Additionally, this research work explores the impact of integrating the user speech acts into the user queries for the conversational agents which leverages document-based knowledge retrieval to improve the quality of the response and effectively evaluate its performance using standard metrics which cross-reference the chat agent response to a pre-existing answer stored somewhere (information from a custom file). The experiments we conducted showed varying performance across the 5 open-source models used across the 11 standard evaluation metrics used. The impacts of this research are twofold: 
\begin{enumerate}
    \item  It exposes the impact of speech acts in discourse analysis and its broad impact on conversational agents.
    \item It illustrates the potential of combining open-source LLMs with a custom-based document retrieval framework to improve Human-Computer Interaction (HCI).
\end{enumerate}
This work paves the way for more intuitive, context-aware, and adaptive conversational systems.

\section{Literature Review}
It has been seen that open-source LLMs such as LLaMA \cite{touvron2023llama}, Mistral \cite{yu2024breakingceilingllmcommunity}, and TinyLLama among others have risen as a relevant and needful alternative to closed-source models because of the scalability, flexibility, and creativity in which they offer. These open-source models have demonstrated imminent relevance in various NLP and machine learning tasks such as natural language generation (NLG) and natural language understanding (NLU). While praising these models for their remarkable ability to capture semantic and syntactic nuances in various tasks, they still struggle with capturing pragmatic nuances such as speech acts \cite{Dresner2003-DREROB-3}. Addressing this limitation opens up a less trivial opportunity for additional linguistic and contextual analysis interfaces to improve conversational applications. 

To give a background of speech acts in Text Analysis, speech acts are often referred to as the communicative intent behind a statement or locution, as defined in speech act theory \cite{austin1962speech} \cite{searle1969speech}. Thus far, there has been a lot of visible significance of the concept of speech acts theoretically, but very limited practical application to NLP. It is seen that more focused text analysis for sentiment, emotion, and intent classification has been enabled by recent improvements in transformer-based architecture models, such as BERT and its derivatives \cite{devlin2018bert}. However, speech act classification remains underexplored and overlooked in computational studies. The core of this study is to build on existing foundations of the speech act theory, leveraging the AIF dataset to model speech acts in dialogue structures, thereby filling in a critical research gap in text analysis for conversational systems.

\section{System Overview}
\begin{figure}[h] 
  \centering
  \includegraphics[width=\linewidth]{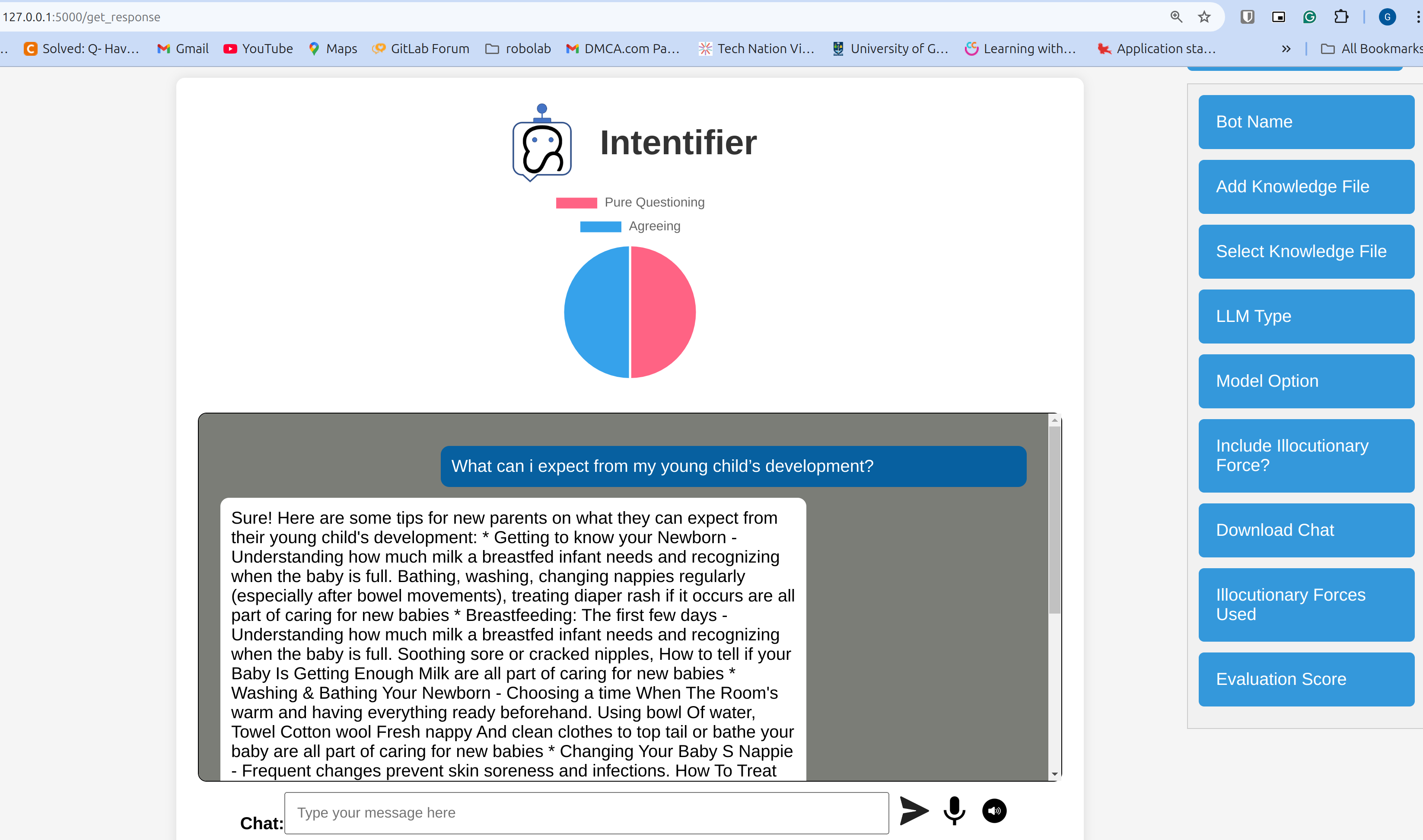}
  \caption{Bespoke Conversational System}
  \label{fig:training}
\end{figure}

The Web Application in Figure \ref{fig:training} was custom-developed for the purpose of this research. It offers the public the ability to carry out research and experiments with speech actss, knowledge base, and open-source LLMs. We give an overview of three key features used for this research below:

\subsection{Model Selection}
This gives end users the ability to choose from a range of various open-source LLM. In the case of this research, we only explored 5 models (llama2:13b Tinyllama:latest Llama3-chatqa-latest Llama3-latest Mistral:latest)

\subsection{Speech Acts Identification}
This component forms the backbone of this research as it gives the ability to either include the user speech acts from their query alongside their query to the conversational agent or just pass their query alone to the conversational agent to generate a response. The speech act identification uses the model stored here \footnote{\href{https://huggingface.co/Godfrey2712/intent_recognition}{Model}}.

\subsection{Document-Based Knowledge Retrieval}
In this integration, we aimed to incorporate speech acts into response generation from information during document retrieval for enhancing contextual alignment between the retrieved information and the user intent. The end users have the ability to upload and select various knowledge files in either .txt or .pdf formats to use as the conversational agent's knowledge to guide its response.

\section{Experiment Design}
To fully understand and expose the impacts of including the speech acts of a user query alongside their query, we set up two key experiments. \\ \\
\textbf{Experiment 1:} In the first experiment, we used two custom documents seen in section 5.1 below as a knowledge base of the model to restrict its knowledge to only the information in the document. We provided two queries for each of the 5 models while assessing without including the users' speech acts, and evaluated the performance of each model based on 10 metrics seen in section 5.5. \\

\begin{center}

\tikzstyle{startstop} = [rectangle, rounded corners, minimum width=0.4cm, minimum height=1cm, text centered, draw=black, fill=red!30]
\tikzstyle{process} = [rectangle, minimum width=1cm, minimum height=1cm, text centered, draw=black, fill=blue!30]
\tikzstyle{arrow} = [thick,->,>=stealth]

\begin{tikzpicture}[node distance=1.8cm]

\node (input) [startstop] {User Query};
\node (agent) [process, below of=input] {Conversational Agent};
\node (model) [process, right of=agent, xshift=4cm] {Ollama Model};
\node (document) [process, left of=agent, xshift=-4cm] {Custom Document};
\node (output) [startstop, below of=agent] {Generated Response};

\draw [arrow] (input) -- (agent);
\draw [arrow] (agent) -- (output);
\draw [arrow] (document) -- (agent);
\draw [arrow] (model) -- (agent);

\end{tikzpicture}
\end{center}

The \textit{User Query} serves as the input to the conversational AI system. This conversational AI system is driven by 5 \textit{Ollama Model} which is guided by \textit{Custom Documents} to produce the required \textit{Generated Response} based on the prompt designed. \\ \\
\textbf{Experiment 2:} In the second experiment, we used two custom documents seen in section 5.1 below as a knowledge base of the model to restrict its knowledge to only the information in the document. We provided two queries and the speech actss extracted from the queries for each of the 5 models and evaluated the performance of each model based on 10 metrics seen in section 5.5. \\

\begin{center}

\tikzstyle{startstop} = [rectangle, rounded corners, minimum width=3cm, minimum height=1cm, text centered, draw=black, fill=red!30]
\tikzstyle{process} = [rectangle, minimum width=0.5cm, minimum height=1cm, text centered, draw=black, fill=blue!30]
\tikzstyle{arrow} = [thick,->,>=stealth]

\begin{tikzpicture}[node distance=1.8cm]

\node (input) [startstop] {User Query + User speech acts};
\node (agent) [process, below of=input] {Conversational Agent};
\node (model) [process, right of=agent, xshift=4cm] {Ollama Model};
\node (document) [process, left of=agent, xshift=-4cm] {Custom Document};
\node (output) [startstop, below of=agent] {Generated Response};

\draw [arrow] (input) -- (agent);
\draw [arrow] (agent) -- (output);
\draw [arrow] (document) -- (agent);
\draw [arrow] (model) -- (agent);

\end{tikzpicture}
\end{center}

The \textit{User Query} and \textit{speech acts} serves as the input to the conversational AI system. This conversational AI system is driven by 5 \textit{Ollama Model} which is guided by \textit{Custom Documents} to produce the required \textit{Generated Response} based on the prompt designed.

\subsection{Chatbot Knowledge Base Data}
\begin{itemize}
    \item Child Health: The Child Health Document is an article from medical experts containing a set of health-related questions and expert answers. \footnote{The article which formed this knowledge file is found at \url{https://www.medicinenet.com/childrens_health/article.htm}}

    \item Poverty in a rising Africa: This is from a World Bank document which looks at the persistent causes of poverty in Africa and the impact on the development of Africa. \footnote{The source document is found here \url{https://www.worldbank.org/content/dam/Worldbank/document/Africa/poverty-rising-africa-poverty-report-main-messages.pdf}}. The full book publication is found here \cite{RePEc:wbk:wbpubs:22575}.
\end{itemize}

\subsection{Prompt Design}
In designing the prompt for the conversational agent, four parameters were used. The `{user\_input}' takes the user query argument from the front-end; The `{context}' consists of a list of the previous conversation between the user and the conversational agent with indicators of users and assistants so that the agent can understand which previous conversation was the user and which was from itself just to ensure coherence; The `{knowledge\_text}' which consist of the text content of the knowledge document selected from the front-end, and the `{illocutionary\_force}' which consist of the speech acts class assigned to the user query after inference from the model stored on hugging-face. The inclusion of the `illocutiomnary\_force' is been controlled from the front-end hence the condition in the prompt which is initialised to `True' but updates based on the choice of the end-user. We show these representations below in a pseudo-code.
\begin{lstlisting}[language=Python]
else:
    # Formulate the updated instruction
    instruction = f"""
    Respond to the user's query: "{user_input}" while considering the relevant context from previous conversations.
    Please focus on providing a response based on the latest exchange, without repeating the entire conversation history.
    If needed, use the context provided below for reference.

    Relevant context: "{context}"

    Use only the information provided in this document to respond: '{knowledge_text}'.        
    """

    if include_illocutionary_force:
        instruction += f"Consider the user's communicative intent while responding, characterized by the speech acts: '{illocutionary_force}'."

    # Call Ollama local model to generate response
    response = ollama.generate(
        model=selected_model,
        prompt=instruction,
        options={
            "temperature": 1,        # No randomness
            "top_p": 1,              # Consider all possible tokens
            "top_k": 1,              # Choose the highest-probability token
            "num_predict": 300,      # Set a fixed length for output
            "seed": 42,              # Ensure reproducibility
            "num_ctx": 4096,         # Maximize context window for tracking history
            "repeat_last_n": -1,      
            "repeat_penalty": 1.5,     
            "mirostat_tau": 1.0   
        },
        stream=False,                # Disable streaming for consistent output
        raw=False                    # Enable prompt formatting for structured input
    )
    response_content = response.get('response', '').strip()
\end{lstlisting}

\subsection{User Queries for each chatbot knowledge base}
\textbf{For Child Health Document (001):} \\\\
\texttt{
\textbf{USER QUERY 1}: What can I expect from my young child’s development? \\
\textbf{USER QUERY 2}: That’s great, thanks for helping.
} \\\\
\textbf{For Poverty in a rising Africa Document (002):} \\\\
\texttt{
\textbf{USER QUERY 1}: It is quite sad to see the poverty level in Africa. What do you think can be done to solve it? \\
\textbf{USER QUERY 2}: Brilliant, that sounds amazing. 
}

\subsection{Conversational agent response evaluation metrics}
\textbf{For Relevance and Quality:}
\begin{itemize}
    \item Recall-Oriented Understudy for Gisting Evaluation (ROUGE-1): Rouge is a conversation metric which is used to evaluate the quality of a response (in most cases - a chatbot) comparing its overlapping unigram with a standardise reference response (normally a human-generated response) \cite{lin-2002-rouge}. In this work, ROUGE-1 is used to measure the presence of certain keywords in the generated response of the chatbot with reference to the custom document used to guide its response. A high ROUGE-1 score indicates that the chatbot's response captures the necessary relevant information from the reference document.
    \item ROUGE-2: Unlike ROUGE-1 which measures unigram with regards to the reference source document, ROUGUE-2 measures the bigram to ensure coherence and contextual alignment between the chatbot response regarding the custom document \cite{see-etal-2017-get}. We can also note that it captures both syntactic and contextual relationships making it ideal for evaluating phrase-level accuracy.
    \item ROUGE-L: ROUGE-L on the other hand evaluates the longest common subsequence between the generated chatbot response and the custom document been used to guide its response. It covers a sentence level, unlike ROUGE-1 and ROUGE-2, making it a good evaluator for syntactic and discourse-level performance. Lin et al., in 2002 \cite{lin-2002-rouge} argued that ROUGE-L was better for capturing sentence level coherence as compared with ROUGE-1 and ROUGE-2. A higher ROUGE-L score indicates that the chatbot response maintains a similar structure as compared to the reference custom document.
    \item Question-Answer Reference (QA-Ref): This measure is used for typical Questions-Answers scenarios where the answer given by the chatbot agent is compared with the answer in the provided knowledge document in the case of this study. QA metrics like the QA-Ref one have been analysed in evaluating performances of conversational agents as seen in Reddy et al. 2019 work \cite{reddy2019coqa}, where QA performance was measured against selected reference answers.
    \item QA-Cand: In scenarios where there could be multiple plausible answers from the reference document, QA-Cand metrics offer a good evaluation standard. It portrays a high tolerance for flexibility in the candidate responses which can be sourced from the custom documents in our case. Candidate-based evaluation has been explored in some open-domain QA tasks, for example in Kwiatkowski et al. 2019 work \cite{kwiatkowski-etal-2019-natural}, where various inclusive answers were considered for evaluating their model performance.

\end{itemize}
\textbf{For Fluency and Coherence:}
\begin{itemize}
    \item BERT-Precision, BERT-Recall, and BERT-F1: These three rely on BERT-embeddings to capture semantic similarities between the chatbot response and the custom document for the case of BERT-Precision, the breadth and coverage of the response based on the custom document for BERT-Recall, and BERT-F1 which gives a balanced tradeoff between the precision (conciseness) and the recall (completeness) offering a more aligned semantic balance.
    \item METEOR: This evaluates precision, recall, stemming, synonyms, and word alignment ensuring a linguistically diverse yet semantically accurate response, with a focus on fluency and flexibility as referenced with the custom document provided. In 2005, Banerjee et al. \cite{banerjee-lavie-2005-meteor} proposed the METEOR as a more linguistically guided alternative to BLEU, developed to offer a better evaluation of language generation tasks.
    \item Perplexity: This generally measures the effect of a probabilistic language model in the ability to predict the following words in a sequence scenario, by calculating the inverse of the probability of the set of test samples which are normalized by the frequency of the words. In 1992, Brown et al. \cite{brown1992class} discussed perplexity as a key metric for evaluating n-gram models while in 2019, Radford et al. \cite{radford2019language} proposed it as a metric for GPT models.

\end{itemize}

\section{Experimental Results and Analysis}
We first installed OLLAMA SDK and pulled the 5 models we needed for our work to our local computer. Our computer had a \textbf{GPU of 12 GB RAM NVIDIA
GEFORCE}. After following the experimental set-up in section 5 above, we recorded the results from all the metrics used from the custom-designed web application including the response time after the query was passed to the model. We show the results in the two tables below with the first one consisting of the results without including the speech acts of the query, and the second table consisting of the results after the speech acts of the query was included. \\
\textbf{Without speech acts added to User Query:} \\\\
\begin{table*}
  \caption{Without speech acts added to User Query:}
  \label{tab:freq}
\resizebox{17cm}{!}{%
  \begin{tabular}{lllllllllllllll}
\toprule
S/N & Documents & Model & Bot & Response & BERT & BERT & BERT & QA & QA & ROUGE-1 & ROUGE-2 & ROUGE-L & METEOR & Perplexity \\
& &  & Response & Time (sec) & Precision & Recall & F1 & Ref & Cand & & &  &  &  \\
\midrule
1 & 001 & llama2:13b & First & 74.64 & 0.72 & 0.72 & 0.72 & 0.49 & 0.24 & 0.24 & 0.13 & 0.19 & 0.42 & 56.75 \\
\textbf{2} & \textbf{001} & \textbf{Tinyllama:latest} & \textbf{First} & \textbf{26.55} & \textbf{0.68} & \textbf{0.68} & \textbf{0.68} & \textbf{0.49} & \textbf{0.39} & \textbf{0.18} & \textbf{0.07} & \textbf{0.12} & \textbf{0.35} & \textbf{26.69} \\
3 & 001 & Llama3-chatqa-latest & First & 74.33 & 0.84 & 0.84 & 0.84 & 0.49 & 0.37 & 0.43 & 0.21 & 0.38 & 0.49 & 54.7 \\
4 & 001 & Llama3-latest & First & 105.02 & 0.74 & 0.74 & 0.74 & 0.49 & 0.63 & 0.16 & 0.01 & 0.09 & 0.22 & 49.16 \\
5 & 001 & Mistral:latest & First & 109.38 & 0.57 & 0.57 & 0.57 & 0.49 & 0.55 & 0.19 & 0.04 & 0.12 & 0.30 & 28.64 \\
6 & 001 & llama2:13b & Second & 164.58 & 0.92 & 0.92 & 0.92 & 0.04 & 0.17 & 0.39 & 0.13 & 0.24 & 0.46 & 70.62 \\
7 & 001 & Tinyllama:latest & Second & 18.04 & 0.81 & 0.81 & 0.81 & 0.04 & 0.09 & 0.33 & 0.09 & 0.18 & 0.41 & 48.82 \\
\textbf{8} & \textbf{001} & \textbf{Llama3-chatqa-latest} & \textbf{Second} & \textbf{62.80} & \textbf{0.90} & \textbf{0.90} & \textbf{0.90} & \textbf{0.04} & \textbf{0.31} & \textbf{0.25} & \textbf{0.08} & \textbf{0.18} & \textbf{0.26} & \textbf{29.12} \\
9 & 001 & Llama3-latest & Second & 85.69 & 0.86 & 0.86 & 0.86 & 0.04 & 0.17 & 0.19 & 0.04 & 0.13 & 0.28 & 55.14 \\
10 & 001 & Mistral:latest & Second & 130.64 & 0.79 & 0.79 & 0.79 & 0.04 & -0.03 & 0.17 & 0.06 & 0.10 & 0.31 & 28.96 \\\\
11 & 002 & llama2:13b & First & 543.83 & 0.78 & 0.78 & 0.78 & 0.77 & 0.63 & 0.12 & 0.05 & 0.10 & 0.21 & 41.60 \\
12 & 002 & Tinyllama:latest & First & 57.02 & 0.84 & 0.84 & 0.84 & 0.77 & 0.03 & 0.10 & 0.00 & 0.08 & 0.15 & 1849.46 \\
13 & 002 & Llama3-chatqa-latest & First & - & - & - & - & - & - & - & - & - & - & - \\
14 & 002 & Llama3-latest & First & 252.28 & 0.74 & 0.74 & 0.74 & 0.77 & 0.68 & 0.11 & 0.04 & 0.08 & 0.19 & 26.80 \\
15 & 002 & Mistral:latest & First & 279.70 & 0.80 & 0.80 & 0.80 & 0.77 & 0.65 & 0.14 & 0.06 & 0.11 & 0.22 & 29.15 \\
16 & 002 & llama2:13b & Second & 532.95 & 0.89 & 0.89 & 0.89 & 0.02 & -0.01 & 0.15 & 0.09 & 0.11 & 0.40 & 45.45 \\
17 & 002 & Tinyllama:latest & Second & 52.84 & 0.93 & 0.93 & 0.93 & 0.02 & 0.03 & 0.19 & 0.05 & 0.19 & 0.20 & 2612.20 \\
18 & 002 & Llama3-chatqa-latest & Second & - & - & - & - & - & - & - & - & - & - & - \\
19 & 002 & Llama3-latest & Second & 237.64 & 0.89 & 0.89 & 0.89 & 0.02 & 0.10 & 0.25 & 0.05 & 0.16 & 0.29 & 35.32 \\
20 & 002 & Mistral:latest & Second & 298.41 & 0.82 & 0.82 & 0.82 & 0.02 & 0.07 & 0.15 & 0.06 & 0.10 & 0.23 & 34.43 \\
\bottomrule
\end{tabular}
}
\end{table*}


\noindent \textbf{With speech acts added to User Query:} \\ \\
\begin{table*}
  \caption{With speech acts added to User Query:}
  \label{tab:freq}
\resizebox{17cm}{!}{%
  \begin{tabular}{lllllllllllllll}
\toprule
S/N & Documents & Model & Bot & Response & BERT & BERT & BERT & QA & QA & ROUGE-1 & ROUGE-2 & ROUGE-L & METEOR & Perplexity \\
& &  & Response & Time (sec) & Precision & Recall & F1 & Ref & Cand & & &  &  &  \\
\midrule
1 & 001 & llama2:13b & First & 212.15 & 0.79 & 0.79 & 0.79 & 0.49 & 0.51 & 0.22 & 0.07 & 0.12 & 0.33 & 72.52 \\
2 & 001 & Tinyllama:latest & First & 24 & 0.74 & 0.74 & 0.74 & 0.49 & 0.48 & 0.23 & 0.10 & 0.13 & 0.41 & 32.90 \\
3 & 001 & Llama3-chatqa-latest & First & 70.92 & 0.64 & 0.64 & 0.64 & 0.49 & 0.68 & 0.24 & 0.02 & 0.14 & 0.24 & 58.74 \\
4 & 001 & Llama3-latest & First & 92.45 & 0.66 & 0.66 & 0.66 & 0.49 & 0.63 & 0.19 & 0.02 & 0.11 & 0.28 & 60.24 \\
5 & 001 & Mistral:latest & First & 107.49 & 0.73 & 0.73 & 0.73 & 0.49 & 0.45 & 0.24 & 0.07 & 0.13 & 0.37 & 28.67 \\
6 & 001 & llama2:13b & Second & 244.50 & 0.79 & 0.79 & 0.79 & 0.043 & 0.066 & 0.22 & 0.07 & 0.12 & 0.33 & 73.07 \\
7 & 001 & Tinyllama:latest & Second & 28.60 & 0.69 & 0.69 & 0.69 & 0.04 & 0.03 & 0.17 & 0.07 & 0.10 & 0.29 & 29.29 \\
8 & 001 & Llama3-chatqa-latest & Second & 64.05 & 0.98 & 0.98 & 0.98 & 0.07 & 0.60 & 0.14 & 0.0 & 0.14 & 0.10 & 70.82 \\
9 & 001 & Llama3-latest & Second & 80.86 & 0.87 & 0.87 & 0.87 & 0.04 & 0.11 & 0.23 & 0.03 & 0.15 & 0.32 & 20.44 \\
10 & 001 & Mistral:latest & Second & 108.78 & 0.83 & 0.83 & 0.83 & 0.04 & 0.09 & 0.25 & 0.06 & 0.17 & 0.37 & 32.75 \\\\
11 & 002 & llama2:13b & First & 611.63 & 0.83 & 0.83 & 0.83 & 0.77 & 0.69 & 0.13 & 0.05 & 0.09 & 0.25 & 39.09 \\
12 & 002 & Tinyllama:latest & First & 59.09 & 0.79 & 0.79 & 0.79 & 0.77 & 0.15 & 0.13 & 0.03 & 0.09 & 0.16 & 2552.25 \\
13 & 002 & Llama3-chatqa-latest & First & - & - & - & - & - & - & - & - & - & - & - \\
14 & 002 & Llama3-latest & First & 350.84 & 0.80 & 0.80 & 0.80 & 0.77 & 0.64 & 0.11 & 0.03 & 0.07 & 0.19 & 31.40 \\
15 & 002 & Mistral:latest & First & 375.14 & 0.76 & 0.76 & 0.76 & 0.77 & 0.61 & 0.12 & 0.04 & 0.09 & 0.28 & 51.88 \\
16 & 002 & llama2:13b & Second & 613.46 & 0.82 & 0.82 & 0.82 & 0.02 & -0.067 & 0.15 & 0.05 & 0.08 & 0.22 & 51.75 \\
17 & 002 & Tinyllama:latest & Second & 67.30 & 0.86 & 0.86 & 0.86 & 0.77 & 0.03 & 0.09 & 0.05 & 0.06 & 0.25 & 583.08 \\
18 & 002 & Llama3-chatqa-latest & Second & - & - & - & - & - & - & - & - & - & - & - \\
19 & 002 & Llama3-latest & Second & 330.94 & 0.86 & 0.86 & 0.86 & 0.019 & 0.017 & 0.21 & 0.042 & 0.12 & 0.28 & 104.75 \\
20 & 002 & Mistral:latest & Second & 329.08 & 0.83 & 0.83 & 0.83 & 0.02 & -0.019 & 0.20 & 0.07 & 0.12 & 0.32 & 41.03 \\
\bottomrule
\end{tabular}
}
\end{table*}

\subsection{Result Analysis}
In the analysis below:\\
F $\Rightarrow$ indicates Faster \\
1 $\Rightarrow$ indicates a better performance \\
0 $\Rightarrow$ indicates a poor performance \\
S $\Rightarrow$ indicates a similar performance \\
\textbf{Analysis of the two results based on first bot responses in Document 001:} \\
\begin{table*}[ht]
\centering
\resizebox{17cm}{!}{%
\begin{tabular}{|l|c|c|c|c|c|c|c|c|c|c|}
\hline
\textbf{} & \multicolumn{5}{c|}{\textbf{Without speech acts}} & \multicolumn{5}{c|}{\textbf{With speech acts}} \\ \hline
\textbf{} & \textbf{llama2} & \textbf{Tinyllama} & \textbf{Llama3-chatqa} & \textbf{Llama3} & \textbf{Mistral} & \textbf{llama2} & \textbf{Tinyllama} & \textbf{Llama3-chatqa} & \textbf{Llama3} & \textbf{Mistral} \\

\textbf{} & \textbf{:13b} & \textbf{:latest} & \textbf{-latest} & \textbf{-latest} & \textbf{:latest} & \textbf{:13b} & \textbf{:latest} & \textbf{-latest} & \textbf{-latest} & \textbf{:latest} \\ \hline

\textbf{Response Time} & F & - & - & - & - & - & F & F & F & F \\ \hline
\textbf{BERT-Precision} & 0 & 0 & 1 & 1 & 0 & 1 & 1 & 0 & 0 & 1 \\ \hline
\textbf{BERT-Recall} & 0 & 0 & 1 & 1 & 0 & 1 & 1 & 0 & 0 & 1 \\ \hline
\textbf{BERT-F1} & 0 & 0 & 1 & 1 & 0 & 1 & 1 & 0 & 0 & 1 \\ \hline
\textbf{QA-Ref} & S & S & S & S & S & S & S & S & S & S \\ \hline
\textbf{QA-Cand} & 0 & 0 & 0 & S & 1 & 1 & 1 & 1 & S & 0 \\ \hline
\textbf{ROUGE-1} & 1 & 0 & 1 & 0 & 0 & 0 & 1 & 0 & 1 & 1 \\ \hline
\textbf{ROUGE-2} & 1 & 0 & 1 & 0 & 0 & 0 & 1 & 0 & 1 & 1 \\ \hline
\textbf{ROUGE-L} & 1 & 0 & 1 & 0 & 0 & 0 & 1 & 0 & 1 & 1 \\ \hline
\textbf{METEOR} & 1 & 0 & 1 & 0 & 0 & 0 & 1 & 0 & 1 & 1 \\ \hline
\textbf{Perplexity} & 1 & 1 & 1 & 1 & 1 & 0 & 0 & 0 & 0 & 0 \\ \hline
\textbf{Avg Total} & 5 & 1 & 8 & 4 & 2 & 4 & 8 & 1 & 4 & 7 \\ \hline
\end{tabular}
}
\caption{Performance metrics for various models for the bot's first response using the first document}
\label{tab:performance}
\end{table*}

From the table above, we analysed the following:
\begin{enumerate}
    \item Response Time: It is observed that the response time of the chat agent was faster in all the models for the experiment which included the speech acts except for the llama2:13b model in which it was faster in the experiment without the speech acts included in the user query.
    \item BERT-Precision: It is observed that the semantic similarities based on the BERT-embeddings were better in Llama3-chatqa-latest and Llama3-latest models in the experiment without including the speech acts, and was better in the rest of the other models in the experiment which included the speech acts.
    \item BERT-Recall: It is observed that the breadth and coverage similarities based on the BERT-embeddings of the chat agent response and the custom document were better in Llama3-chatqa-latest and Llama3-latest models in the experiment without including the speech acts, and was better in the rest of the other models in the experiment which included the speech acts.
    \item BERT-F1: It is observed that a balanced trade-off between the precision and recall which gives a better semantic balance based on the BERT-embeddings of the chat agent response and the custom document were better in Llama3-chatqa-latest and Llama3-latest models in the experiment without including the speech acts, and was better in the rest of the other models in the experiment which included the speech acts.
    \item QA-Ref: Comparing the chat agent response typically with answers in the knowledge document, it is observed that all the models in both the experiment without and with the speech acts included had the same performance indicating a balanced metric.
    \item QA-Cand: When we observed the flexibility of the response with respect to the custom document provided, it was seen that llama2:13b, Tinyllama:latest, Llama3-chatqa-latest performed better in the experiment with the speech acts included, performed same with llama3-latest, and performed better in mistral:latest for the experiment without the speech acts.
    \item ROUGE-1: It is observed that elements of unigram keywords in the chat agent response found in the custom document were better in llama2:13b and Llama3-chatqa-lates models in the experiment without including the speech acts, and were better in the rest of the other models in the experiment which included the speech acts.
    \item ROUGE-2: It is observed that elements of Bigram indicating coherence in the chat agent response based on the custom document were better in llama2:13b and Llama3-chatqa-lates models in the experiment without including the speech acts, and were better in the rest of the other models in the experiment which included the speech acts.
    \item ROUGE-L: It is observed that the longest common subsequence between the chat agent response and the custom document was better in llama2:13b and Llama3-chatqa-latest models in the experiment without including the speech acts, and was better in the rest of the other models in the experiment which included the speech acts.
    \item METEOR: It is observed that a better linguistically diverse yet semantically response from the chat agent with respect to the custom document, was better in llama2:13b and Llama3-chatqa-latest models for the experiment conducted without including the speech acts, and better in the other models for the experiment conducted with the inclusion of the speech acts with the user query.
    \item Perplexity: The ability to predict the following words in a sequence scenario was better in all the models for the experiment conducted without including the speech acts with a lower perplexity value as compared to same models for the experiment conducted with the inclusion of the speech acts with the user query.
    \item Avg Total: Taking a record of the performance of the models across all metrics and excluding both the response time and the results of the similar values, it is observed that: the llama2:13b model performs better with an average score of 5 in the experiment without the speech acts against same model for the experiment with the speech acts which had a score of 4; the Tinyllama:latest model performs better with an average score of 8 in the experiment with the speech acts against same model for the experiment without the speech acts which had a score of 1; the Llama3-chatqa-latest model performs better with an average score of 8 in the experiment without the speech acts against same model for the experiment with the speech acts which had a score of 1; the Llama3-latest model had similar score of 4 in the experiment with the speech acts and for the experiment without the speech acts which also had a score of 4; the Mistral:latest model performs better with an average score of 7 in the experiment with the speech acts against same model for the experiment without the speech acts which had a score of 2.
    
\end{enumerate}
\textbf{Analysis of the two results based on second bot responses in Document 001:}\\
\begin{table*}[ht]
\centering
\resizebox{17cm}{!}{%
\begin{tabular}{|l|c|c|c|c|c|c|c|c|c|c|}
\hline
\textbf{} & \multicolumn{5}{c|}{\textbf{Without speech acts}} & \multicolumn{5}{c|}{\textbf{With speech acts}} \\ \hline
\textbf{} & \textbf{llama2} & \textbf{Tinyllama} & \textbf{Llama3-chatqa} & \textbf{Llama3} & \textbf{Mistral} & \textbf{llama2} & \textbf{Tinyllama} & \textbf{Llama3-chatqa} & \textbf{Llama3} & \textbf{Mistral} \\

\textbf{} & \textbf{:13b} & \textbf{:latest} & \textbf{-latest} & \textbf{-latest} & \textbf{:latest} & \textbf{:13b} & \textbf{:latest} & \textbf{-latest} & \textbf{-latest} & \textbf{:latest} \\ \hline
\textbf{Response Time} & F & F & F & - & - & - & - & - & F & F \\ \hline
\textbf{BERT-Precision} & 1 & 1 & 0 & 0 & 0 & 0 & 0 & 1 & 1 & 1 \\ \hline
\textbf{BERT-Recall} & 1 & 1 & 0 & 0 & 0 & 0 & 0 & 1 & 1 & 1 \\ \hline
\textbf{BERT-F1} & 1 & 1 & 0 & 0 & 0 & 0 & 0 & 1 & 1 & 1 \\ \hline
\textbf{QA-Ref} & 0 & S & 0 & S & S & 1 & S & 1 & S & S \\ \hline
\textbf{QA-Cand} & 1 & 1 & 0 & 1 & 0 & 0 & 0 & 1 & 0 & 1 \\ \hline
\textbf{ROUGE-1} & 1 & 1 & 1 & 0 & 0 & 0 & 0 & 0 & 1 & 1 \\ \hline
\textbf{ROUGE-2} & 1 & 1 & 1 & 1 & S & 0 & 0 & 0 & 0 & S \\ \hline
\textbf{ROUGE-L} & 1 & 1 & 1 & 0 & 0 & 0 & 0 & 0 & 1 & 1 \\ \hline
\textbf{METEOR} & 1 & 1 & 1 & 0 & 0 & 0 & 0 & 0 & 1 & 1 \\ \hline
\textbf{Perplexity} & 1 & 0 & 1 & 0 & 1 & 0 & 1 & 0 & 1 & 0 \\ \hline
\textbf{Avg Total} & 9 & 8 & 5 & 2 & 1 & 1 & 1 & 5 & 7 & 7 \\ \hline
\end{tabular}
}
\caption{Performance metrics for various models for the bot's second response using the first document}
\label{tab:performance}
\end{table*}

From the table above, we analysed the following:
\begin{enumerate}
    \item Response Time: It is observed that the response time of the chat agent was faster in Llama3-latest and  Mistral:latest for the experiment which included the speech acts and was faster for the other models in the experiment without the speech acts included in the user query.
    \item BERT-Precision: It is observed that the semantic similarities based on the BERT-embeddings were better in llama2:13b and Tinyllama:latest models in the experiment without including the speech acts, and were better in the rest of the other models in the experiment which included the speech acts.
    \item BERT-Recall: It is observed that the breadth and coverage similarities based on the BERT-embeddings of the chat agent response and the custom document were better in llama2:13b and Tinyllama:latest models in the experiment without including the speech acts, and was better in the rest of the other models in the experiment which included the speech acts.
    \item BERT-F1: It is observed that a balanced trade-off between the precision and recall which gives a better semantic balance based on the BERT-embeddings of the chat agent response and the custom document were better in llama2:13b and Tinyllama:latest models in the experiment without including the speech acts, and was better in the rest of the other models in the experiment which included the speech acts.
    \item QA-Ref: Comparing the chat agent response typically with answers in the knowledge document, it is observed that Tinyllama:latest, Llama3-latest, and Mistral:latest models in both the experiment without and with the speech acts included had the same performance indicating a balanced metric, but was better llama2:13b, and Llama3-chatqa-latest in the experiment with the inclusion of the speech acts as compared to the experiment without the speech acts.
    \item QA-Cand: When we observed the flexibility of the response with respect to the custom document provided, it was seen that Llama3-chatqa-latest, and Mistral:latest performed better in the experiment with the speech acts included, and performed better in other models for the experiment without the speech acts.
    \item ROUGE-1: It is observed that elements of unigram keywords in the chat agent response found in the custom document were better in llama2:13b, Tinyllama:latest, and Llama3-chatqa-latest models in the experiment without including the speech acts, and were better in the rest of the other models in the experiment which included the speech acts.
    \item ROUGE-2: It is observed that elements of bigram indicating coherence in the chat agent response based on the custom document were better in all the models in the experiment without including the speech acts, and was same in only the Mistral:latest model in the experiment which included the speech acts.
    \item ROUGE-L: It is observed that the longest common subsequence between the chat agent response and on the custom document was better in llama2:13b, Tinyllama:latest, and Llama3-chatqa-latest  models in the experiment without including the speech acts, and were better in the rest of the other models in the experiment which included the speech acts.
    \item METEOR: It is observed the a better linguistically diverse but yet sementically response from the chat agent with respect to the custom document, was better in llama2:13b, Tinyllama:latest, and Llama3-chatqa-latest models for the experiment conducted without including the speech acts, and better in the other models for the experiment conducted with the inclusion of the speech acts with the user query.
    \item Perplexity: The ability to predict the following words in a sequence scenario was better in llama2:13b, Llama3-chatqa-latest, and Mistral:latest models for the experiment conducted without including the speech acts with a lower perplexity value and better in other models for the experiment conducted with the inclusion of the speech acts with the user query.
    \item Avg Total: Taking a record of the performance of the models across all metrics and excluding both the response time and the results of the similar values, it is observed that: the llama2:13b model performs better with an average score of 9 in the experiment without the speech acts against same model for the experiment with the speech acts which had a score of 1; the Tinyllama:latest model performs better with an average score of 8 in the experiment without the speech acts against same model for the experiment with the speech acts which had a score of 1; the Llama3-chatqa-latest model performs same with an average score of 5 in the experiment without the speech acts against same model for the experiment with the speech acts which had a similar score of 5; the Llama3-latest model performed better with a score of 7 in the experiment with the speech acts and for the experiment without the speech acts with a score of 2; the Mistral:latest model performs better with an average score of 7 in the experiment with the speech acts against same model for the experiment without the speech acts which had a score of 1.
\end{enumerate}
\textbf{Analysis of the two results based on first bot responses in Document 002}\\
\begin{table*}[ht]
\centering
\resizebox{17cm}{!}{%
\begin{tabular}{|l|c|c|c|c|c|c|c|c|c|c|}
\hline
\textbf{} & \multicolumn{5}{c|}{\textbf{Without speech acts}} & \multicolumn{5}{c|}{\textbf{With speech acts}} \\ \hline
\textbf{} & \textbf{llama2} & \textbf{Tinyllama} & \textbf{Llama3-chatqa} & \textbf{Llama3} & \textbf{Mistral} & \textbf{llama2} & \textbf{Tinyllama} & \textbf{Llama3-chatqa} & \textbf{Llama3} & \textbf{Mistral} \\

\textbf{} & \textbf{:13b} & \textbf{:latest} & \textbf{-latest} & \textbf{-latest} & \textbf{:latest} & \textbf{:13b} & \textbf{:latest} & \textbf{-latest} & \textbf{-latest} & \textbf{:latest} \\ \hline
\textbf{Response Time} & F & F & - & F & F & - & - & - & - & - \\ \hline
\textbf{BERT-Precision} & 0 & 1 & - & 0 & 1 & 1 & 0 & - & 1 & 0 \\ \hline
\textbf{BERT-Recall} & 0 & 1 & - & 0 & 1 & 1 & 0 & - & 1 & 0 \\ \hline
\textbf{BERT-F1} & 0 & 1 & - & 0 & 1 & 1 & 0 & - & 1 & 0 \\ \hline
\textbf{QA-Ref} & S & S & - & S & S & S & S & - & S & S \\ \hline
\textbf{QA-Cand} & 0 & 0 & - & 1 & 1 & 1 & 1 & - & 0 & 0 \\ \hline
\textbf{ROUGE-1} & 0 & 0 & - & S & 1 & 1 & 1 & - & S & 0 \\ \hline
\textbf{ROUGE-2} & S & 0 & - & 1 & 1 & S & 1 & - & 0 & 0 \\ \hline
\textbf{ROUGE-L} & 1 & 0 & - & 1 & 1 & 0 & 1 & - & 0 & 0 \\ \hline
\textbf{METEOR} & 0 & 0 & - & S & 0 & 1 & 1 & - & S & 1 \\ \hline
\textbf{Perplexity} & 0 & 1 & - & 1 & 1 & 1 & 0 & - & 0 & 0 \\ \hline
\textbf{Avg Total} & 1 & 4 & - & 4 & 8 & 7 & 5 & - & 3 & 1 \\ \hline
\end{tabular}
}
\caption{Performance metrics for various models for the bot first response using the second document}
\label{tab:performance}
\end{table*}

From the table above, we analysed the following:
\begin{enumerate}
    \item Response Time: It is observed that the response time for Llama2:13b, Tinyllama:latest Llama3-latest, and Mistral:latest is faster in the experiment conducted without the speech acts as compared with the same models in the experiment with the speech acts.

    \item BERT-Precision: It is observed that the precision metric based on BERT-embeddings showed a better score for Tinyllama:latest and Mistral:latest in the experiment conducted without including the speech acts and a better score was observed in the Llama3-latest, and Llama2:13b models for the experiment that included the speech acts in the user query.

    \item BERT-Recall: It is observed that the recall metric based on BERT-embeddings is similar to the precision results with better scores noted for Tinyllama:latest and Mistral:latest in the experiment without including the speech acts, and better scores for Llama3-latest and Llama2:13b in the experiment with the speech acts included with the user query.

    \item BERT-F1: It is observed that the balanced precision and recall showed similar results to BERT-Recall and Precision, indicating a better score for Tinyllama:latest and Mistral:latest in the experiment without speech acts, and better scores for Llama3-latest and Llama2:13b with the speech acts included with the user query.

    \item QA-Ref: Comparing the chat agent response typically with answers in the knowledge document, it is observed that the performance was similar across all models in both experiments, indicating a high similarity in this metric.

    \item QA-Cand: When we observed the flexibility of the response with respect to the custom document provided, it was seen that Llama3-latest and Mistral:latest showed better results in the experiment without the speech acts, and better results for Llama2:13b and Tinyllama:latest in the experiment when the speech acts was included with the user query.

    \item ROUGE-1: It is observed that elements of unigram keywords in the chat agent response found in the custom document were better for Mistral:latest in the experiment without the speech acts, while Llama2:13b and Tinyllama:latest models showed better scores in the experiment with the speech acts included, and a similar score for llama3-latest in both experiments.

    \item ROUGE-2: It is observed that bigram coherence was better for Llama3-latest and Mistral:latest in the experiment without the speech acts, but was observed as similar for Llama2:13b across both experiments and better in Tinyllama:latest with the inclusion of the speech acts.

    \item ROUGE-L: It is observed that the longest common subsequence metric was better for Llama2:13b, Llama3-latest, and Mistral:latest in the experiment conducted without the speech acts. However, Tinyllama:latest was the only model to achieve a better result in the experiment with the speech acts included in the user query.

    \item METEOR: It is observed that linguistic diversity and semantic accuracy in responses showed a better score for Llama2:13b, Tinyllama:latest, and Mistral:latest in the experiment with the speech acts, but was observed to be similar for Llama3-latest in both experiments.

    \item Perplexity: It is observed that the predictive ability of the models indicated better values for Tinyllama:latest, Llama3-latest and Mistral:latest in the experiment without the speech acts, while the same metric achieved a better score in Llama2:13b for the experiment with the speech acts.

    \item Avg Total: Taking a record of the performance of the models across all metrics, it was observed that the Mistral:latest model performed best with a total score of 8 in the experiment without the speech acts and a much lower score of 1 in the experiment with the speech acts. Similarly, Tinyllama:latest obtained a score of 4 in the experiment without the speech acts but increased to 5 when the speech acts was included. Llama2:13b had a significant difference in performance, with a score of 1 in the experiment without speech acts but improved to 7 when it was included in the user query. Finally, Llama3-latest showed moderate performance, scoring 4 without the speech acts and 3 when it was included in the user query.
\end{enumerate}

\textbf{Analysis of the two results based on second bot responses in Document 002}

\begin{table*}[ht]
\centering
\resizebox{17cm}{!}{%
\begin{tabular}{|l|c|c|c|c|c|c|c|c|c|c|}
\hline
\textbf{} & \multicolumn{5}{c|}{\textbf{Without speech acts}} & \multicolumn{5}{c|}{\textbf{With speech acts}} \\ \hline
\textbf{} & \textbf{llama2} & \textbf{Tinyllama} & \textbf{Llama3-chatqa} & \textbf{Llama3} & \textbf{Mistral} & \textbf{llama2} & \textbf{Tinyllama} & \textbf{Llama3-chatqa} & \textbf{Llama3} & \textbf{Mistral} \\

\textbf{} & \textbf{:13b} & \textbf{:latest} & \textbf{-latest} & \textbf{-latest} & \textbf{:latest} & \textbf{:13b} & \textbf{:latest} & \textbf{-latest} & \textbf{-latest} & \textbf{:latest} \\ \hline
\textbf{Response Time} & F & F & - & F & F & - & - & - & - & - \\ \hline
\textbf{BERT-Precision} & 1 & 1 & - & 1 & 0 & 0 & 0 & - & 0 & 1 \\ \hline
\textbf{BERT-Recall} & 1 & 1 & - & 1 & 0 & 0 & 0 & - & 0 & 1 \\ \hline
\textbf{BERT-F1} & 1 & 1 & - & 1 & 0 & 0 & 0 & - & 0 & 1 \\ \hline
\textbf{QA-Ref} & S & 0 & - & 1 & S & S & 1 & - & 0 & S \\ \hline
\textbf{QA-Cand} & 1 & S & - & 1 & 1 & 0 & S & - & 0 & 0 \\ \hline
\textbf{ROUGE-1} & S & 1 & - & 1 & 0 & S & 0 & - & 0 & 1 \\ \hline
\textbf{ROUGE-2} & 1 & S & - & 1 & 0 & 0 & S & - & 0 & 1 \\ \hline
\textbf{ROUGE-L} & 1 & 1 & - & 1 & 0 & 0 & 0 & - & 0 & 1 \\ \hline
\textbf{METEOR} & 1 & 0 & - & 1 & 0 & 0 & 1 & - & 0 & 1 \\ \hline
\textbf{Perplexity} & 1 & 0 & - & 1 & 1 & 0 & 1 & - & 0 & 0 \\ \hline
\textbf{Avg Total} & 8 & 5 & - & 10 & 2 & 0 & 3 & - & 0 & 7 \\ \hline
\end{tabular}
}
\caption{Performance metrics for various models for the bot's second response using the second document}
\label{tab:performance}
\end{table*}

From the table above, we analysed the following:
\begin{enumerate}
    \item Response Time: It is observed that the response time for Llama2:13b, Tinyllama:latest Llama3-latest, and Mistral:latest is faster in the experiment conducted without the speech acts as compared with the same models in the experiment with the speech acts.

    \item BERT-Precision: It is observed that the precision metric based on BERT-embeddings showed a better score for llama2:13b, Tinyllama:latest and Llama3-latest in the experiment conducted without including the speech acts and a better score was observed in Mistral:latest model for the experiment that included the speech acts in the user query.

    \item BERT-Recall: It is observed that the recall metric based on BERT-embeddings is similar to the precision results with better scores for llama2:13b, Tinyllama:latest and Llama3-latest in the experiment conducted without including the speech acts and a better score was observed in Mistral:latest model for the experiment that included the speech acts in the user query.

    \item BERT-F1: It is observed that the balanced precision and recall showed similar results to BERT-Recall and Precision, indicating better scores for llama2:13b, Tinyllama:latest and Llama3-latest in the experiment conducted without including the speech acts and a better score was observed in Mistral:latest model for the experiment that included the speech acts in the user query.

    \item QA-Ref: Comparing the chat agent response typically with answers in the knowledge document, it is observed that the performance was similar for llama2:13b and Mistral:latest in both experiments, indicating a high similarity in this metric but Llama3-latest was better in the first experiment, and Tinyllama:latest better in the second experiment.

    \item QA-Cand: When we observed the flexibility of the response with respect to the custom document provided, it was seen that Llama2:13b, Llama3-latest and Mistral:latest showed better results in the experiment without the speech acts, and the same result for Tinyllama:latest in both experiments.

    \item ROUGE-1: It is observed that elements of unigram keywords in the chat agent response found in the custom document were better for Tinyllama:latest, and llama3-latest in the experiment without the speech acts, while Mistral:latest model showed better scores in the experiment with the speech acts included, and a similar score for Llama2:13b in both experiments.

    \item ROUGE-2: It is observed that bigram coherence was better for Llama2:13b and Llama3-latest in the experiment without the speech acts, but was observed as similar for Tinyllama:latest across both experiments and better in Mistral:latest with the inclusion of the speech acts.

    \item ROUGE-L: It is observed that the longest common subsequence metric was better for Llama2:13b, Llama3-latest, and Tinyllama:lates in the experiment conducted without the speech acts. However, Mistral:latest was the only model to achieve a better result in the experiment with the speech acts included in the user query.

    \item METEOR: It is observed that linguistic diversity and semantic accuracy in responses showed a better score for Llama2:13b, and Llama3-latest in the experiment with the speech acts, but was observed to be better for Tinyllama:latest, and Mistral:latest in the experiment with the speech acts included in the user query.

    \item Perplexity: It is observed that the predictive ability of the models indicated better values for Llama2:13b, Llama3-latest and Mistral:latest in the experiment without the speech acts, while the same metric achieved a better score in Tinyllama:latest for the experiment with the speech acts.

    \item Avg Total: Taking a record of the performance of the models across all metrics, it was observed that the Llama3-latest model performed best with a total score of 10 in the experiment without the speech acts and a much lower score of 0 in the experiment with the speech acts. Similarly, llama2:13b obtained a score of 8 in the experiment without the speech acts but dropped to 0 when the speech acts was included. Tinyllama:latest showed a difference in performance, with a score of 5 in the experiment without speech acts but dropped to 3 when it was included in the user query. Finally, Mistral:latest showed a performance of 2 without the speech acts and was improved to 7 when it was included in the user query.
\end{enumerate}

\section{Conclusion and Future work}
The results from this research unlock several potentials for the inclusion of communicative intent (speech acts) which have been overlooked by computational linguistics in improving the responses given by conversational agents in an HCI setting. The speech actss explored in this research originate from a dialogue domain making it very relevant as an additional linguistic feature layer which is not currently embedded in the context of transformer models. To improve on this work, there is a need to explore larger encoder-decoder models with larger parameters and explore the impact of repeating these experiments on a more robust GPU. The open-source can be found here \footnote{\href{https://github.com/Godfrey2712/convostem}{ConvoStem}}.

\newpage
\bibliographystyle{acm}  
\bibliography{main.bib}  

\begin{thebibliography}{10}

\bibitem{austin1962speech}
{\sc Austin, J.}
\newblock Speech acts, 1962.

\bibitem{banerjee-lavie-2005-meteor}
{\sc Banerjee, S., and Lavie, A.}
\newblock {METEOR}: An automatic metric for {MT} evaluation with improved correlation with human judgments.
\newblock In {\em Proceedings of the {ACL} Workshop on Intrinsic and Extrinsic Evaluation Measures for Machine Translation and/or Summarization\/} (Ann Arbor, Michigan, June 2005), J.~Goldstein, A.~Lavie, C.-Y. Lin, and C.~Voss, Eds., Association for Computational Linguistics, pp.~65--72.

\bibitem{RePEc:wbk:wbpubs:22575}
{\sc Beegle, K., Christiaensen, L., Dabalen, A., and Gaddis, I.}
\newblock {\em {Poverty in a Rising Africa}}.
\newblock No.~22575 in World Bank Publications - Books. The World Bank Group, 2016.

\bibitem{brown1992class}
{\sc Brown, P.~F., Della~Pietra, V.~J., Desouza, P.~V., Lai, J.~C., and Mercer, R.~L.}
\newblock Class-based n-gram models of natural language.
\newblock {\em Computational linguistics 18}, 4 (1992), 467--480.

\bibitem{devlin2018bert}
{\sc Devlin, J.}
\newblock Bert: Pre-training of deep bidirectional transformers for language understanding.
\newblock {\em arXiv preprint arXiv:1810.04805\/} (2018).

\bibitem{Dresner2003-DREROB-3}
{\sc Dresner, E.}
\newblock Review of bunt \& black (2000): Abduction, belief and context in dialogue--studies in computational pragmatics.
\newblock {\em Pragmatics and Cognition 11}, 2 (2003), 390--394.

\bibitem{kwiatkowski-etal-2019-natural}
{\sc Kwiatkowski, T., Palomaki, J., Redfield, O., Collins, M., Parikh, A., Alberti, C., Epstein, D., Polosukhin, I., Devlin, J., Lee, K., Toutanova, K., Jones, L., Kelcey, M., Chang, M.-W., Dai, A.~M., Uszkoreit, J., Le, Q., and Petrov, S.}
\newblock Natural questions: A benchmark for question answering research.
\newblock {\em Transactions of the Association for Computational Linguistics 7\/} (2019), 452--466.

\bibitem{radford2019language}
{\sc Radford, A., Wu, J., Child, R., Luan, D., Amodei, D., Sutskever, I., et~al.}
\newblock Language models are unsupervised multitask learners.
\newblock {\em OpenAI blog 1}, 8 (2019), 9.

\bibitem{reddy2019coqa}
{\sc Reddy, S., Chen, D., and Manning, C.~D.}
\newblock Coqa: A conversational question answering challenge.
\newblock {\em Transactions of the Association for Computational Linguistics 7\/} (2019), 249--266.

\bibitem{searle1969speech}
{\sc Searle, J.}
\newblock {\em Speech Acts: An Essay in the Philosophy of Language}.
\newblock Cam: Verschiedene Aufl. Cambridge University Press, 1969.

\bibitem{see-etal-2017-get}
{\sc See, A., Liu, P.~J., and Manning, C.~D.}
\newblock Get to the point: Summarization with pointer-generator networks.
\newblock In {\em Proceedings of the 55th Annual Meeting of the Association for Computational Linguistics (Volume 1: Long Papers)\/} (Vancouver, Canada, July 2017), R.~Barzilay and M.-Y. Kan, Eds., Association for Computational Linguistics, pp.~1073--1083.

\bibitem{touvron2023llama}
{\sc Touvron, H., Lavril, T., Izacard, G., Martinet, X., Lachaux, M.-A., Lacroix, T., Rozi{\`e}re, B., Goyal, N., Hambro, E., Azhar, F., et~al.}
\newblock Llama: Open and efficient foundation language models.
\newblock {\em arXiv preprint arXiv:2302.13971\/} (2023).

\bibitem{yu2024breakingceilingllmcommunity}
{\sc Yu, Y.-C., Kuo, C.-C., Ye, Z., Chang, Y.-C., and Li, Y.-S.}
\newblock Breaking the ceiling of the llm community by treating token generation as a classification for ensembling, 2024.

\end{thebibliography}

\end{document}